\documentclass[10pt,twocolumn,letterpaper]{article}

\usepackage{cvpr}            
\usepackage[accsupp]{axessibility}

\usepackage{graphicx}
\usepackage{amsmath}
\usepackage{amssymb}
\usepackage{booktabs}
\usepackage{multirow}
\usepackage[normalem]{ulem}
\useunder{\uline}{\ul}{}
\usepackage{fancyhdr}

\usepackage[capitalize]{cleveref}
\crefname{section}{Sec.}{Secs.}
\Crefname{section}{Section}{Sections}
\Crefname{table}{Table}{Tables}
\crefname{table}{Tab.}{Tabs.}

\def\confName{CVPR}
\def\confYear{2023}

\begin{document}

\title{Multi-Granularity Archaeological Dating of Chinese Bronze Dings Based on a Knowledge-Guided Relation Graph}

\author{
Rixin Zhou\textsuperscript{1} \qquad
Jiafu Wei\textsuperscript{1} \qquad
Qian Zhang\textsuperscript{3} \qquad
Ruihua Qi\textsuperscript{3} \qquad
Xi Yang\textsuperscript{1,2,}\textsuperscript{*} \qquad
Chuntao Li\textsuperscript{3,}\textsuperscript{*}\\
\textsuperscript{1}School of Artificial Intelligence, Jilin University\\
\textsuperscript{2}Engineering Research Center of Knowledge-Driven Human-Machine Intelligence, MoE, China \\
\textsuperscript{3}School of Archaeology, Jilin University\\
}




\twocolumn[{
\renewcommand\twocolumn[1][]{#1}
\maketitle
\begin{center}
    \captionsetup{type=figure}
    \centering
    \includegraphics[width=1.0\linewidth]{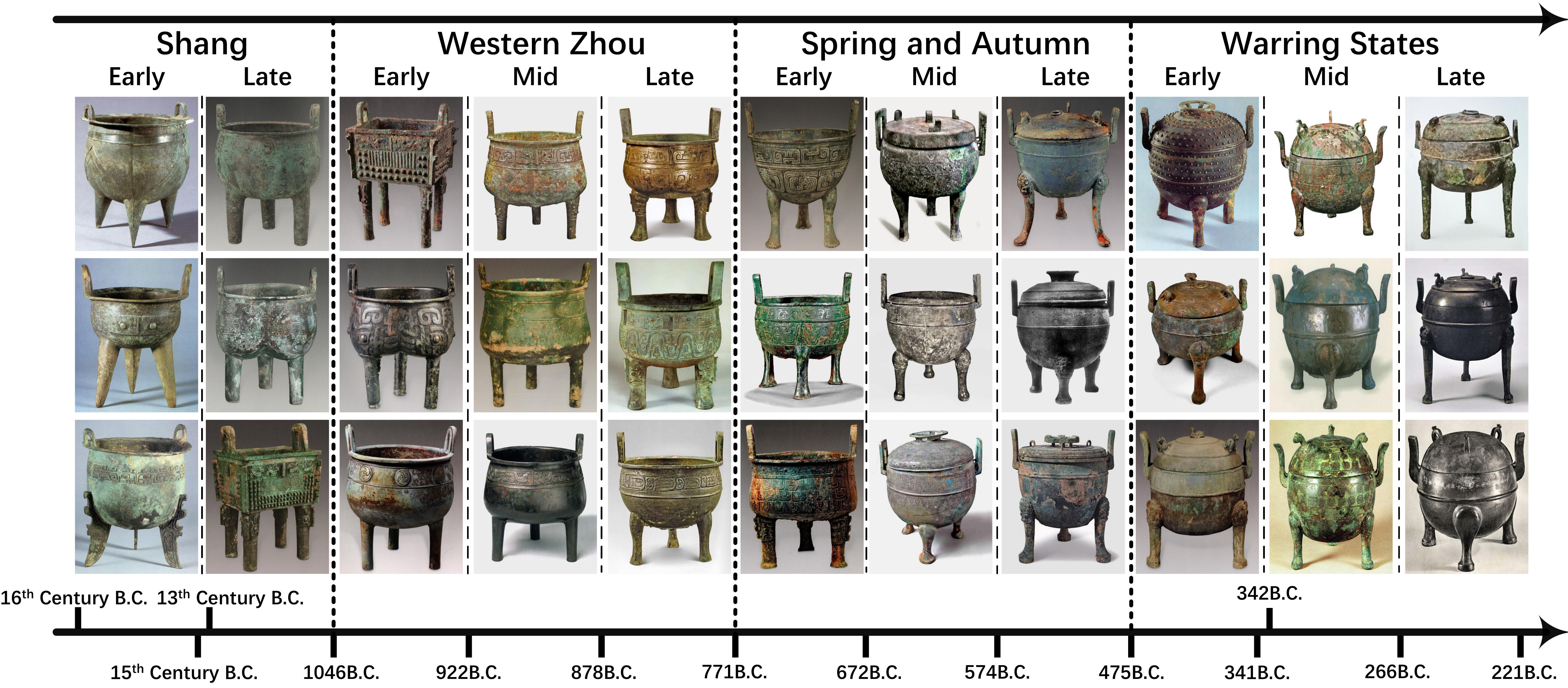}
    \captionof{figure}{Typical examples of Chinese bronze dings from 4 dynasties and 11 periods. The columns from left to right show Shang (Early, Late), Western Zhou (Early, Mid, Late), Spring and Autumn (Early, Mid, Late), and Warring States (Early, Mid, Late). The timeline under the image is the time range for the corresponding periods, where B.C. indicates Before Christ.}
  \label{fig1:examples}
\end{center}

}]
\begin{abstract}
   \vspace{-0.5cm}
   The archaeological dating of bronze dings has played a critical role in the study of ancient Chinese history. Current archaeology depends on trained experts to carry out bronze dating, which is time-consuming and labor-intensive. For such dating, in this study, we propose a learning-based approach to integrate advanced deep learning techniques and archaeological knowledge. To achieve this, we first collect a large-scale image dataset of bronze dings, which contains richer attribute information than other existing fine-grained datasets. Second, we introduce a multihead classifier and a knowledge-guided relation graph to mine the relationship between attributes and the ding era. Third, we conduct comparison experiments with various existing methods, the results of which show that our dating method achieves a state-of-the-art performance. We hope that our data and applied networks will enrich fine-grained classification research relevant to other interdisciplinary areas of expertise. The dataset and source code used are included in our supplementary materials, and will be open after submission owing to the anonymity policy. Source codes and data are available at: https://github.com/zhourixin/bronze-Ding.
\end{abstract}
\thispagestyle{fancy} 
      \fancyhead{} 
      \fancyfoot{} 
      \fancyfoot[L]{\footnotesize *Corresponding authors}
      \renewcommand{\headrulewidth}{0pt} 
      \renewcommand{\footrulewidth}{1pt} 
      \fancyfootoffset[R]{-14cm}

\vspace{-10.5pt}
\section{Introduction}
\vspace{-4pt}
Dings are cauldrons used for cooking, storage, and ritual offerings to gods or ancestors in ancient China, and they are the most important species used in Chinese ritual bronzes~\cite{49}. The archaeological dating of dings has contributed to the study of ancient Chinese history. Although the excavated bronzes are massive, dating such artifacts depends on the long-term training and accumulation of expertise in archaeological typology~\cite{54}. In addition, some artifacts are easy to identify to a precise age and others are difficult to identify.


For the object, we focus on a ding, the features of which are similar and complicated in different eras, as shown in the columns of Figure~\ref{fig1:examples}. We therefore consider this dating task as a fine-grained classification problem. Simultaneously, research into fine-grained classification is close to that of other areas of expertise because it often requires expensive specialized data and knowledge areas, such as birds (zoology)~\cite{12,56} and flowers (botany)~\cite{55}.

Data features and domain knowledge, particularly in archaeology, vary in different fields. In addition to the common traits of the existing fine-grained datasets, our data are more challenging. First, our data are unbalanced and difficult to mitigate through their collection because they are determined based on an unearthed state. Second, there are more similarities between bronze dings of adjacent eras, leading to the possibility of misclassifying them into fine granularity adjacent eras beyond a coarse granularity. In other words, compared to other fine-grained classification data, our data have a larger intra-class difference and a smaller inter-class difference between adjacent eras. Third, the attributes and eras are intertwined and the relations are more complex. Each period of bronze dings has multiple shapes and characteristics, and each shape and characteristic correspond to multiple periods of bronze dings, leading to the impracticality of making simple judgments regarding the period based on the shape and characteristic. Existing fine-grained classification methods therefore struggle when applying our data.

To address these issues, we make the following contributions in this study:
\begin{itemize}
\setlength{\itemsep}{0pt}
\setlength{\parsep}{0pt}
\setlength{\parskip}{0pt}
\item We collect an image dataset of 3690 bronze dings with rich annotations made by bronze experts, including the era (4 course-grained dynasties and 11 fine-grained periods), attributes (29 shapes and 96 characteristics with bounding boxes), literature, location of excavation, and the museum where they are displayed. 
\item We build an end-to-end multihead network to solve this multi-granularity task. The two heads combine coarse- and fine-grained features in a bidirectional manner with a gradient truncated addition to improve the performance at both granularities. The outputs of other two heads, the shape and characteristic nodes, are added to a knowledge-guided relation graph to embed the domain knowledge into our network,  
\item We propose exploiting these rich attributes following archaeological knowledge by employing the focal-type probability classification loss and indicate the ineffectiveness of simply concatenating external information. 
\item We achieve the best performance in terms of the dating accuracy, outperforming other state-of-the-art (SOTA) fine-grained classification methods. 
\end{itemize}

\section{Related Work}
\vspace{-4pt}
\subsection{Bronze Dings dating}
\vspace{-4pt}

In addition to manual inference~\cite{54, 28}, the chemical and physical properties of metals are also used to locate the exact year of a bronze ding~\cite{10,11,23, 24}. However, chemical and material science based techniques are time-consuming and difficult to manipulate, and they may cause irreversible damage to the bronze. Meanwhile, machine-learning-based methods have been used to explore bronze inscription recognition~\cite{15,21,22}. Although the meaning of the inscriptions on a bronze ding is important, it is insufficient for achieving an accurate dating. The automatic dating of bronze dings using artificial intelligence has been largely underexplored.

\subsection{Fine-Grained Visual Classification}
\paragraph{Datasets.}
\vspace{-4pt}
Compared to a traditional image recognition task~\cite{3,4,5}, a fine-grained visual classification (FGVC) task is more challenging~\cite{44}. Although the variation between different categories of fine-grained data can be extremely small, the variation within the same category, owing to changes in pose and occlusions, is much broader, which leads to more difficulties. Several datasets have been proposed to address these challenges, including birds~\cite{12,56,57}, dogs~\cite{14}, airplanes~\cite{13}, flowers~\cite{55}, cars~\cite{58}, vegetables~\cite{59}, fruits~\cite{59}, foods~\cite{60}, fashion~\cite{61,62,41}, and retail products~\cite{63,64}.

\vspace{-10.5pt}

\paragraph{Single-Granularity Visual Classification.}
Single-granularity FGVC treats objects at the single-class level. Some studies~\cite{6,7,8,9,45,47} have been based on localization-classification networks to find the local features with differentiation and then combine the global features for fine-grained recognition. In addition, high-order feature interactions~\cite{29,30,31} and the design of specific loss functions~\cite{32,33,34,35,48} have resulted in significant improvements. In addition to conventional methods, to further assist fine-grained recognition tasks, some researchers leverage external information, such as attributes ~\cite{46,66,67,68,69}, web data~\cite{36,70,71}, multi-modal data~\cite{37,72,73}, or human-computer interactions~\cite{38,74}.
\vspace{-10.5pt}

\paragraph{Multi-Granularity Visual Classification.}
Hierarchical multi-granularity structures can express richer information than single-granularity structures. We construct the network as a hierarchical structure, which has also been adapted in a number of other studies~\cite{39,40,1,20,2}. In protein function prediction, the output of each level is combined with the input of the next level, thus allowing the network to learn the features of each level jointly~\cite{39}. Parameters assigned for each task are used to encourage cross-task feature interactions~\cite{40}. Tree-structured tasks are constructed to integrate knowledge from the tree hierarchy~\cite{20,43} and conduct a feature transfer between levels~\cite{2,65}.

\section{Bronze Ding Dataset}
\label{section3}
\vspace{-4pt}
\paragraph{Data Collection and annotation.}
\label{sec:data Collect and Anno}
We collect more than four thousand ding images from both five published archaeology books and four websites, and sort out 3690 images as our dataset. Some of these are line graphs. Because many dating results are controversial, we re-argue the era of each artifact through discussions with three bronze experts. The collection and labelling of data were carried out by an archaeologist and eight archaeology assists, who took approximately 8 months to complete.

\begin{figure}[t]
\centering
  \includegraphics[width=0.98\linewidth]{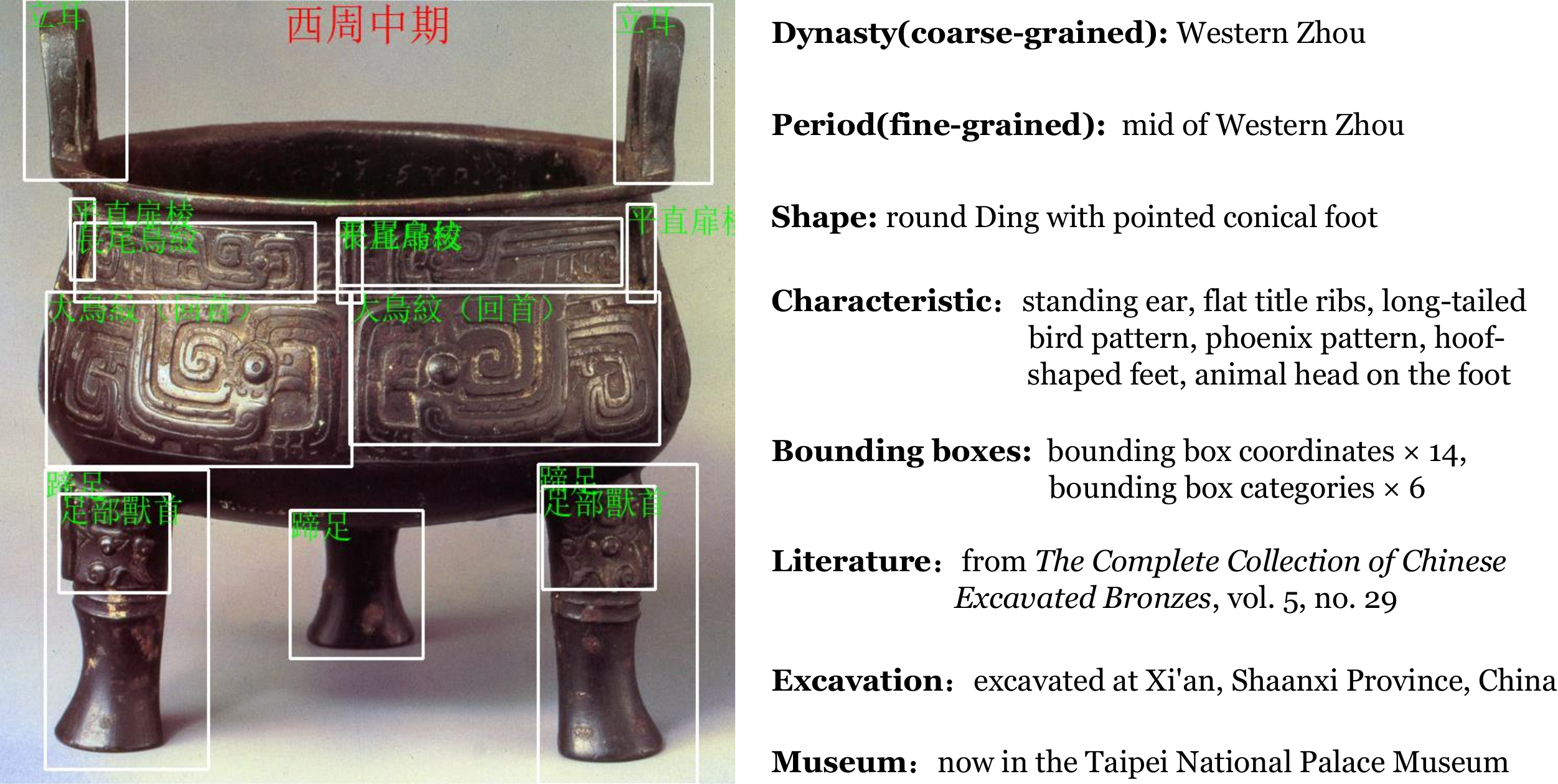}
  \caption{Annotations of a bronze ding example. The characteristics are labeled by bounding boxes in the left figure, and the right shows the detailed information.}
  \label{fig3:bbox}
\end{figure}

The collected dings belong to 4 course-grained dynasties and 11 fine-grained periods, and each image is annotated with additional annotations, as shown in Figure~\ref{fig3:bbox}, including:

\begin{itemize}
\setlength{\itemsep}{0pt}
\setlength{\parsep}{0pt}
\setlength{\parskip}{0pt}
\item \textbf{Shape}: Single-category label for bronze ding shape, with 29 types in total.
\item \textbf{Characteristic}: Multi-category labels for the key components, decorations, and inscriptions, along with bounding boxes, with 96 types in total.
\item \textbf{Source}: Literature (studies to which these bronze ding images belong); excavation (location of excavation), and museum (current exhibition museums).
\end{itemize}
The detailed process used in the data annotation is described in the supplementary material.

\begin{figure}[htbp]
\centering
  \includegraphics[width=0.98\linewidth]{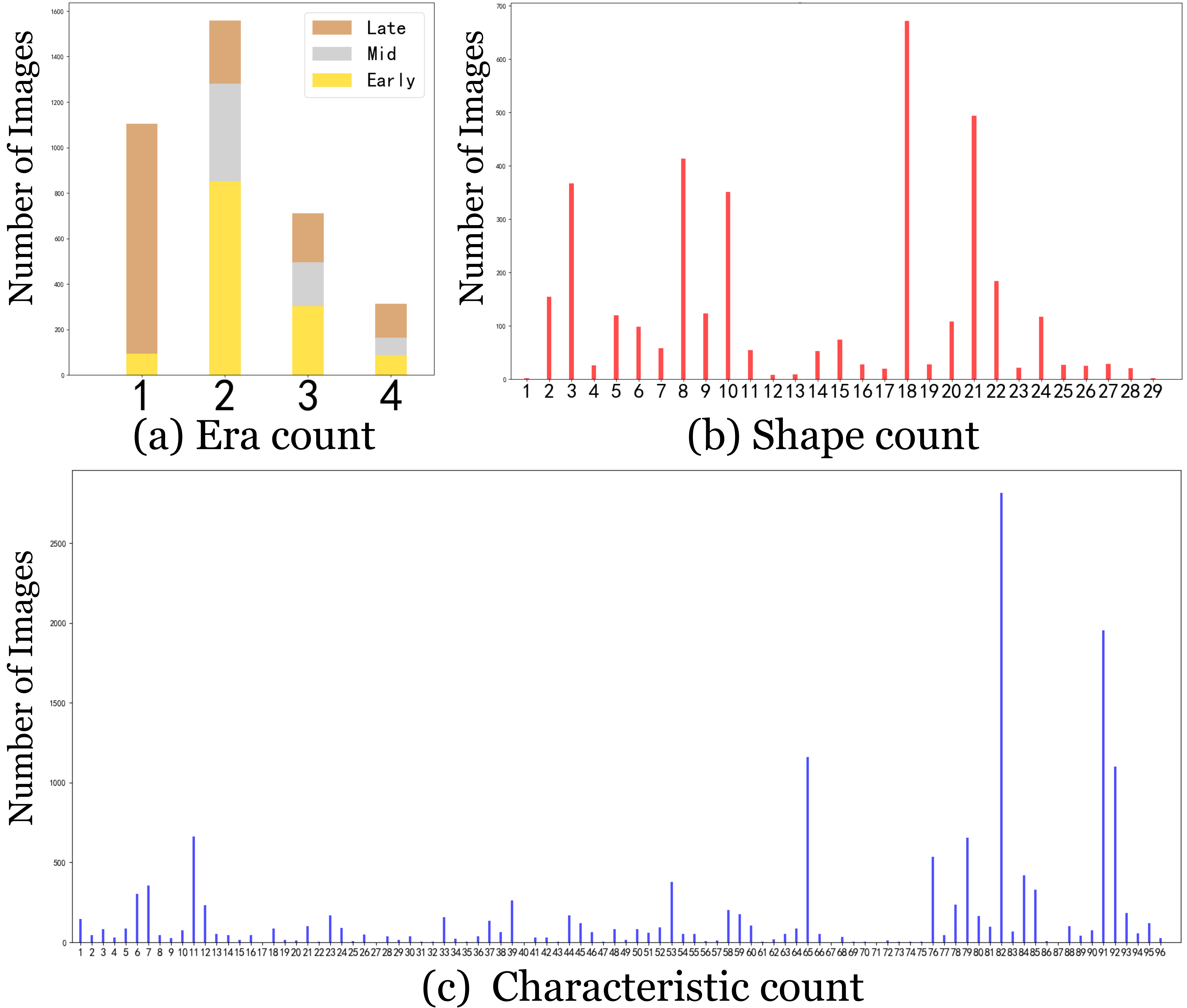}
  \caption{Statistics showing the imbalance of our dataset.}
  \label{fig4:bar}
\end{figure}

 \vspace{-10.5pt}
\paragraph{Statistics.}
We also count the numbers of eras, shapes, and characteristic labels from different eras, the results of which are shown in Figure~\ref{fig4:bar}. The numbers of shapes and characteristic annotations varied considerably between eras.

\begin{table}[t]
\centering
\caption{Comparison of our dataset and five existing datasets in terms of category statistics, entropy, conditional entropy, and information gain. The information gain shows that the shape and characteristic annotations of our dataset provide richer information.}
\label{tab2:entropy}
\resizebox{0.48\textwidth}{!}{  
\begin{tabular}{c|cc|c|cc|c|c|c}
\hline\hline
                                      & \multicolumn{2}{c|}{Bronze Ding}                                                                                                                                  & CUB\_200\_2011         & \multicolumn{2}{c|}{Deep Fashion}                                                                                                      & CompCars
                                      & Stanford Dogs           & Food-101                \\ \hline\hline
Images                                & \multicolumn{2}{c|}{3690}                                                                                                                                         & 11788                  & \multicolumn{2}{c|}{289222}                                                                                                            & 136726                         & 20580                   & 101000                  \\ \hline
Image Categories                      & \multicolumn{2}{c|}{11}                                                                                                                                           & 200                    & \multicolumn{2}{c|}{50}                                                                                                                & 1716                           & 120                     & 101                     \\ \hline
Images per Category                   & \multicolumn{2}{c|}{$335\pm315$}                                                                                                                     & $59\pm3$  & \multicolumn{2}{c|}{$5784\pm11989$}                                                                                       & $79\pm48 $        & $171\pm23$ & $1000\pm0$ \\ \hline
\multirow{2}{*}{Attribute Categories} & \multicolumn{1}{c|}{96}                                                                        & 29                                                               & \multirow{2}{*}{312}   & \multicolumn{1}{c|}{1000}                                                     & 26                                                     & \multirow{2}{*}{218}           & \multirow{2}{*}{\textbf{$--$}}      & \multirow{2}{*}{\textbf{$--$}}      \\
                                      & \multicolumn{1}{c|}{$(characteristic)$}                                                          & $(shape)$                                                          &                        & \multicolumn{1}{c|}{$(coarse)$}                                                 & $(fine)$                                                 &                                &                         &                         \\ \hline
$H\left(D\right)$                                & \multicolumn{2}{c|}{3.459}                                                                                                                                        & 7.644                  & \multicolumn{2}{c|}{5.644}                                                                                                             & 10.745                         & \textbf{$--$}                       & \textbf{$--$}                       \\ \hline
\multirow{2}{*}{$H\left(D \mid A\right)$}        & \multicolumn{1}{c|}{1.826}                                                                     & 1.717                                                            & \multirow{2}{*}{6.599} & \multicolumn{1}{c|}{4.470}                                                    & 4.789                                                  & \multirow{2}{*}{9.570}         & \multirow{2}{*}{\textbf{$--$}}      & \multirow{2}{*}{\textbf{$--$}}      \\
                                      & \multicolumn{1}{c|}{$(characteristic)$}                                                          & $(shape)$                                                          &                        & \multicolumn{1}{c|}{$(coarse)$}                                                 & $(fine)$                                                 &                                &                         &                         \\ \hline
$g(D, A)$                             & \multicolumn{1}{c|}{\textbf{\begin{tabular}[c]{@{}c@{}}1.633\\ $(characteristic)$\end{tabular}}} & \textbf{\begin{tabular}[c]{@{}c@{}}1.742\\ $(shape)$\end{tabular}} & 1.045                  & \multicolumn{1}{c|}{\begin{tabular}[c]{@{}c@{}}1.174\\ $(coarse)$\end{tabular}} & \begin{tabular}[c]{@{}c@{}}0.855\\ $(fine)$\end{tabular} & 1.002                          & \textbf{$--$}                       & \textbf{$--$}                       \\ \hline
\end{tabular}
}
\end{table}

\paragraph{Comparison.}
To quantify the information provided by the additional annotations, we calculate the information gain of the shape and characteristic annotations of the era judgement, as shown in Equation (\ref{eq1:entropy}).

\begin{equation}
g(D, A)=H(D)-H(D \mid A)
\label{eq1:entropy}
\end{equation}
where $\mathrm{H}(\mathrm{D})$ is the entropy of the fine-grained labels on dataset $D$ and $\mathrm{H}(\mathrm{D}\mid\mathrm{A})$ is the conditional entropy of attribute $A$ on dataset $D$.

For comparison, the entropy in CUB-200-2011~\cite{14} and Deep-Fashion~\cite{41} are also calculated. The results are presented in Table~\ref{tab2:entropy}. We find that the information gain in our dataset is more significant than that of the other datasets, which means that our shape and characteristic annotations provide richer information. Such information is therefore critical for improving the network performance.

\begin{figure*}[htbp]
\centering
  \includegraphics[width=1.0\linewidth]{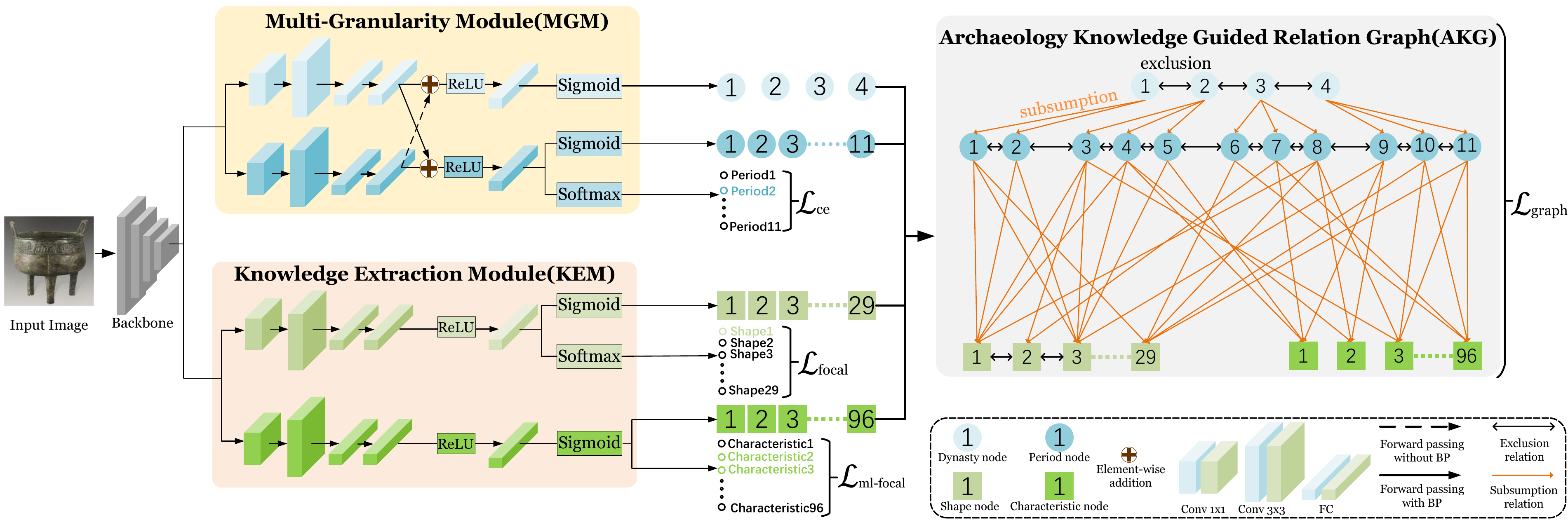}
  \caption{Overview of our network. We design four heads: two heads in MGM are responsible for extracting dynasty and period features at two granularities. Two other heads in KEM are responsible for extracting shape and characteristic features of the bronze dings. Then, the outputs of MGM and KEM jointly build our AKG to formulate the relationship between the eras and attributes of each ding through the graph loss. Simultaneously, the outputs are also used to compute cross-entropy and focal losses to enhance the learning of the annotations on each head.}
  \label{fig2:architecture}
\end{figure*}

\section{Methodology}
\vspace{-4pt}
\subsection{Overview}
\vspace{-4pt}
To address the challenges to our task, we construct a multihead network for predicting the era of bronze dings on an archaeology knowledge-guided relation graph, as shown in Figure ~\ref{fig2:architecture}. The network consists of three parts: a multi-granularity module (MGM), knowledge extraction module (KEM), and archaeology knowledge-guided relation graph (AKG). Compared to HRN~\cite{2}, first, we additionally enhance the dynasty (coarse) features without affecting the period (fine) dating performance by adding the feature of period head to the dynasty head with gradient truncation, and then feeding into the next FC layer. Second, we implement KEM in our network and extend the relation graph to leverage the features of our dataset by considering attributes information (shape and characteristic) in the knowledge graph. Third, we design a focal-type probability classification loss to learn the relationship between attributes and eras from easy (shape) to difficult (characteristic), instead of just learning category information.

\subsection{Network Architecture}
\vspace{-4pt}
\paragraph{Multi-granularity Module.}

The MGM is built to combine dynasty features with period features and enhance them interactively. After the backbone network, dynasty and period features are separately extracted by two heads consisting of two convolution layers and two fully connected layers. Then, in addition to applying an element-wise addition from dynasty head to period head to enrich period information, an element-wise gradient truncation addition is applied in reverse to enhance dynasty features without affecting the period performance.



For the outputs, dynasty head applies a $sigmoid$ projection, and then the result forms the dynasty node of the AKG. The period head applies $sigmoid$ and $softmax$ projections, where the $sigmoid$ output forms the period node of the AKG, and the $softmax$ output computes the cross-entropy loss $\mathcal{L}_{ce}$ to enhance the exclusive relation between the period nodes.

\vspace{-10.5pt}
\paragraph{Knowledge Extraction Module.}
Because of the narrow inter-class differences and wide intra-class variations of the bronze ding, archaeologists must combine various factors to determine the era to which they belong. We therefore introduce domain knowledge to improve the performance of the network, including the shape and characteristic annotations. Specifically, we design a KEM for extracting the shape and characteristic information of a bronze ding, which consists of two separate heads for extracting knowledge-specific features.

The shape head of the KEM applies $sigmoid$ and $softmax$ projections. From Figures~\ref{fig4:bar} (b) and (c), we can see that the shape and characteristic categories are unbalanced; therefore, we compute $\mathcal{L}_{\text {focal}}$~\cite{42} between its $softmax$ output and the shape labels to alleviate the category imbalance problem and enhance the exclusive relation between the shape nodes. Furthermore, because the classification of the characteristics is a multilabel classification task, the characteristic head of the KEM only applies a $sigmoid$ projection, and we compute the multilabel $\mathcal{L}_{\text {ml-focal}}$ between its $sigmoid$ output and the characteristic labels. The $sigmoid$ outputs of these two heads are also fed to the AKG.

\subsection{Archaeology Knowledge Guided Relation Graph}
\vspace{-4pt}
\paragraph{The Formalism of Relation Graph.}
Inspired by the study in~\cite{2,20}, we develop an archaeological knowledge-guided relation graph embedded with domain knowledge to enable the network to synthetically learn the era, shape, and characteristic labels. The nodes of the relation graph are the types of eras and attributes from MGM and KEM, and a set of directed edges and undirected edges are defined between these nodes. A directed edge is a subsumption edge that indicates that the parent nodes subsume the child node. An undirected edge is an exclusion edge and indicates that the two nodes are mutually exclusive.

According to archaeological knowledge, we conclude that the relations of the edges and nodes are as follows:
\begin{itemize}
\setlength{\itemsep}{0pt}
\setlength{\parsep}{0pt}
\setlength{\parskip}{0pt}
\item Because a bronze ding cannot belong to two eras at the same time, any two dynasty or period nodes only have an exclusive edge between them. The relation of the era node subsumes in the dynasty node and can be expressed as a subsumption edge.
\item Although one bronze ding only contains one type of shape node, multiple dings within the same period contain multiple shape nodes. Therefore, the period and shape nodes have multiple subsumption edges. In addition, because a bronze ding cannot have two shapes, any two period nodes have exclusive edges between them.
\item A bronze ding may contain multiple characteristics, and the period and characteristic nodes therefore have multiple subsumption edges.
\end{itemize}
Based on these relations, we define an extended legal global assignment of all labels in the relation graph as binary-label vectors for an object. Beyond~\cite{2}, we consider the shape and characteristics of a node assignment. The set of all legal global assignments forms the era state space $S_{G_e} \subseteq \left\{0, 1\right\}^n$, era-shape combination state space $S_{G_{es}} \subseteq \left\{0, 1\right\}^{n+m}$, and era-characteristic combination state space $S_{G_{ec}} \subseteq \left\{0, 1\right\}^{n+k}$ of relation graph $G$, where $n$, $m$, and $k$ denote the number of nodes for the eras, shapes, and characteristics, respectively. Thus, we calculate the probabilistic classification loss on $G$, enabling the network to improve its judgement of the era by learning domain knowledge of the shapes and characteristics.

\vspace{-10.5pt}

\paragraph{Focal-type Probabilistic Classification Loss.} 
The two types of attributes have different effects on the dating by archaeologists. There are a few types of shapes, which are relatively easy to distinguish. However, the ding shape type is not decisive for dating. Meanwhile, some characteristic types can accurately define this era. However, the number of characteristic types is large, and they are both similar and complex. Based on this knowledge, we construct probabilistic classification losses to enable the network to learn information in the relation graph. 

During training, we obtain the predicted label in the relation graph and maximized its marginal probability in a step-by-step manner. Given an input image $\mathbf{x}$, the unnormalized era joint probability of all era nodes concerning the era label assignment $\mathbf{y_e}$ can be computed as $\tilde{P}_{e}(\mathbf{y_{e}} \!\!\mid\!\!  \mathbf{x})$.
The era joint probability is then normalized by $\operatorname{Pr_{e}}(\mathbf{y_{e}} \mid \mathbf{x})=\frac{\tilde{P_{e}}(\mathbf{y_{e}} \mid \mathbf{x})}{Z_e(\mathbf{x})}$, where $Z_e(\mathbf{x})$ is the era partition function that sums over all legal era assignments $\overline{\mathbf{y}}_e \in S_{G_e}$ in the era state space. If input image $\mathbf{x}$ has the $i$-th era label, we can obtain the era marginal probability $\operatorname{Pr_{e}}(y_{e_i}=1 \mid \mathbf{x})$ of era label $i$ by summing over all legal era assignments $\overline{\mathbf{y}}_e$ that include $\overline{y}_{e_i}$ = 1. Procedures for calculating normalized era-shape joint probability $\operatorname{Pr_{es}}(\mathbf{y}_{es} \mid \mathbf{x})$ and era-characteristic joint probability$\operatorname{Pr_{ec}}(\mathbf{y}_{ec} \mid \mathbf{x})$ are the same. The details of the calculations are described in the supplementary material.

Given $m$ training samples, $\mathcal{D}=\left\{x^{l}, y_e^{l}, y_{es}^{l}, y_{ec}^{l}, g_{e}^{l}, \notag\right. \\ \left. g_{es}^{l}, g_{ec}^{l}\,\right\}$, $l=1, \ldots, m$, where $y_e^{l}$, $y_{es}^{l}$ and $y_{ec}^{l}$ are the ground-truth label vector of the era, era-shape combination, and era-characteristic combination, respectively. And $g_{e}^{l}  \in 
 \left\{1,\ldots , n\right\} $, $g_{es}^{l}\in \left\{1,\ldots , n+m\right\}$, $g_{ec}^{l}\in \left\{1,\ldots , n+k\right\}$ are the indices of the observed era, era-shape combination, and era-characteristic combination labels, respectively. Subsequently, the era probabilistic classification loss $\mathcal{L}_{\text {e}}(\mathcal{\!D\!})$ is defined as follows:
 \begin{equation}
-\frac{1}{m} \sum_{l=1}^{m} \ln (\operatorname{Pr_e}(y_{e_{g_{e} ^{l}}}^{l}=1 \mid \mathbf{x}^{l})\!)
\label{eq:Lp}
\end{equation}
 

Then, because of the different importance of attributes, we define the focal-type era-shape probabilistic classification loss $\mathcal{L}_{\text {es}}(\mathcal{\!D\!})$ and the era-shape-characteristic probabilistic classification loss $\mathcal{L}_{\text {esc}}(\mathcal{\!D\!})$ as follows:
\begin{equation}
-\frac{1}{m} \!\!\displaystyle\sum_{l=1}^{m}\!\!\left(\!\!(\!1\!-\!\operatorname{Pr_{e}}\!(y_{e_{g_{e} ^{l}}}^{l}\!\!\!=\!1 \!\!\mid \!\!\mathbf{x}^{l})\!)^{	\alpha_1} \! \ln  (\operatorname{Pr_{es}}(y_{{es}_{g_{es} ^{l}}}^{l}\!\!\!\!\!\!=\!\!1 \!\!\mid \!\!\mathbf{x}^{l})\!)\!\!\right)
\label{eq:Le-s}
\end{equation}

\vspace{-10.5pt}

\begin{equation}
-\frac{1}{m} \!\!\displaystyle\sum_{l=1}^{m}\!\!\left(\!\!(\!1\!\!-\!\!\operatorname{Pr_{es}}\!(y_{{es}_{g_{es} ^{l}}}^{l}\!\!\!\!\!\!=\!1 \!\!\mid \!\!\mathbf{x}^{l})\!)^{	\alpha_2} \! \ln  (\!\operatorname{Pr_{ec}}(y_{{ec}_{g_{ec} ^{l}}}^{l}\!\!\!\!\!=\!\!1 \!\!\mid \!\!\mathbf{x}^{l})\!)\!\!\right)
\label{eq:Le-a}
\end{equation}

Thus, when the network learns sufficient information through the era features of a given sample to determine its era, the influence of the shape and characteristics can be weakened by decay factor $\alpha_1$ and $\alpha_2$. When the network cannot learn a sufficient amount of information to determine the era of this sample, $\mathcal{L}_{\text {es}}(\mathcal{\!D\!})$ plays a supportive role. When neither the era features nor the shape features can provide sufficient information, $\mathcal{L}_{\text {esc}}(\mathcal{\!D\!})$ will contribute to determining the chronology of this sample by learning the relationship between the era and characteristics. In this manner, the network can adaptively adjust its learning of the shape and characteristics according to the amount of information possessed by different samples, thereby avoiding a disturbance of the main task.

Finally, the aforementioned losses are added in a linear manner to form a complete probabilistic classification loss:
\begin{equation}
\mathcal{L}_{\text {graph}}(\mathcal{D})= \mathcal{L}_{\text {e}}(\mathcal{D}) + \beta \cdot (\mathcal{L}_{\text {es}}(\mathcal{D}) + \mathcal{L}_{\text {esc}}(\mathcal{D}))
\label{eq:graph}
\end{equation}
where $\beta$ denotes the weight used to balance the influence of the loss components. 

\subsection{Total Loss}
\vspace{-4pt}
In addition to the probabilistic classification loss $\mathcal{L}_{graph}$, we also use $\mathcal{L}_{ce}$, $\mathcal{L}_{focal}$, and $\mathcal{L}_{ml-focal}$ to enable the network to learn the era, shape, and characteristic categories of bronze dings, respectively. 
In summary, the total loss is given as a summation of the aforementioned losses with a trade-off parameter, $\lambda$:
\begin{equation}
\mathcal{L}_{\text {total }}(\!\mathcal{D}\!)\!=\!\mathcal{L}_{\text {graph }}(\!\mathcal{D}\!) \!+\! \mathcal{L}_{\text {ce}}(\!\mathcal{D}\!)\! +\! \lambda \!\cdot \!(\mathcal{L}_{\text {focal}}(\!\mathcal{D}\!) \!+\! \mathcal{L}_{\text {ml-focal}}(\!\mathcal{D}\!))
\label{eq:all}
\end{equation}


\section{Experiments}
\vspace{-4pt}
\subsection{Data Preparation}
\vspace{-4pt}
We split our data into three sets: the scales of the training set, validation set, and test set are 4:1:5 (1470:363:1857), respectively, following the divisions of other fine-grained classification datasets~\cite{12,13}. Due to the data imbalance, we keep the proportions of each dynasty and period divided in the same way. For the pre-processing, we apply data augmentation to the collected images, including background removal and grayscale.

\subsection{Implementation Details}
\label{ImDetails}
\vspace{-4pt}
We implement our network using PyTorch~\cite{51} and conduct experiments on a workstation equipped with an NVIDIA RTX 3090 GPU. For a fair comparison, we also adopt ResNet50 pretrained on ImageNet as our network backbone and resized the input images to 400$\times$400 throughout the experiments. We train each experiment for 64 epochs with early stopping and use Adam optimizer with a learning rate of 0.0001, adjusted using a cosine annealing strategy~\cite{27} to optimize our network. The batch size is set to 32. Besides, We set the decay factors $\alpha_1=2$ and $\alpha_2=3$ in Equations (\ref{eq:Le-s}) and (\ref{eq:Le-a}), balance weight $\beta=0.001$ in Equation (\ref{eq:graph}), and trade-off parameter $\lambda=0.1$ in Equation (\ref{eq:all}). And the parameter settings in $\mathcal{L}_{focal}$ and $\mathcal{L}_{ml-focal}$ follow~\cite{42}. The impact of hyper-parameters are also analysed in supplementary materials. Based on these implementation details, we set the HRN~\cite{2} as our baseline model.

\vspace{-10.5pt}

\paragraph{Evaluation.} 
We use the overall accuracy ($OA$) and the area under the average precision and recall curve $AU(\overline{PRC})$ to evaluate the dating performance.

\begin{table} 
\centering
\caption{Ablation study of each component in our network.} 
\label{tab:abalation}
\resizebox{.45\textwidth}{!}{  
\begin{tabular}{c|c|cc|cccc|cc}
\hline\hline
\multirow{3}{*}{} & \multirow{3}{*}{\begin{tabular}[c]{@{}c@{}}MGM \\ w/ Truncated\end{tabular}} & \multicolumn{2}{c|}{KEM}                                 & \multicolumn{4}{c|}{AKG}                                                               & \multirow{3}{*}{Dynasty $OA$} & \multirow{3}{*}{Period $OA$} \\ \cline{3-8}
                  &                                                                              & \multirow{2}{*}{Shape} & \multirow{2}{*}{Characteristic} & \multicolumn{2}{c|}{Shape}                       & \multicolumn{2}{c|}{Characteristic} &                             &                            \\ \cline{5-8}
                  &                                                                              &                        &                                 & Concat       & \multicolumn{1}{c|}{Embed}    & Concat           & Embed        &                             &                            \\ \hline\hline
1                 &                                                                              &                        &                                 &              & \multicolumn{1}{c|}{}             &                  &                  & 85.28                       & 75.81                      \\ \hline
2                 & $\checkmark$                                                                 &                        &                                 &              & \multicolumn{1}{c|}{}             &                  &                  & 86.85                       & 77.05                      \\ \hline
3                 & $\checkmark$                                                                 & $\checkmark$           &                                 &              & \multicolumn{1}{c|}{}             &                  &                  & 84.54                       & 75.54                      \\ \hline
4                 & $\checkmark$                                                                 & $\checkmark$           &                                 & $\checkmark$ & \multicolumn{1}{c|}{}             &                  &                  & 86.53                       & 76.51                      \\ \hline
5                 & $\checkmark$                                                                 & $\checkmark$           &                                 &              & \multicolumn{1}{c|}{$\checkmark$} &                  &                  & 87.71                       & 77.86                      \\ \hline
6                 & $\checkmark$                                                                 &                        & $\checkmark$                    &              & \multicolumn{1}{c|}{}             &                  &                  & 87.23                       & 77.37                      \\ \hline
7                 & $\checkmark$                                                                 &                        & $\checkmark$                    &              & \multicolumn{1}{c|}{}             & $\checkmark$     &                  & 87.50                       & 77.10                      \\ \hline
8                 & $\checkmark$                                                                 &                        & $\checkmark$                    &              & \multicolumn{1}{c|}{}             &                  & $\checkmark$     & 87.72                       & 77.98                      \\ \hline
9                 & $\checkmark$                                                                 & $\checkmark$           & $\checkmark$                    & $\checkmark$ & \multicolumn{1}{c|}{}             & $\checkmark$     &                  & 86.80                       & 76.62                      \\ \hline
10                & $\checkmark$                                                                 & $\checkmark$           & $\checkmark$                    &              & \multicolumn{1}{c|}{$\checkmark$} &                  &                  & 87.12                       & 77.05                      \\ \hline
11                &                                                                              & $\checkmark$           & $\checkmark$                    &              & \multicolumn{1}{c|}{$\checkmark$} &                  & $\checkmark$     & 87.45                       & \textbf{78.83}             \\ \hline
12                & $\checkmark$                                                                 & $\checkmark$           & $\checkmark$                    &              & \multicolumn{1}{c|}{$\checkmark$} &                  & $\checkmark$     & \textbf{88.79}              & \textbf{78.83}             \\ \hline
\end{tabular}
} 
\end{table}

\subsection{Ablation Study}
\vspace{-4pt}
We evaluate different combinations of the proposed components and followed the default parameter settings described in Section ~\ref{ImDetails}. The results are listed in Table ~\ref{tab:abalation}.
\vspace{-10.5pt}
\paragraph{Multi-Granularity Module (MGM).} 
We apply an element-wise gradient truncated addition, from the period features to the dynasty features, forming a bi-directional interaction structure. Simple but efficient, this (1$\rightarrow$2) improves the dynasty dating accuracy of the network by $1.57\%$ and the period dating accuracy by $1.24\%$. The removal of the gradient truncated addition from the complete model (11$\rightarrow$12) leads to a $1.34\%$ decrease in the dynasty dating accuracy, while maintaining the period dating accuracy. 
This result is also in line with our original intention, which strengthens the dynasty dating while not affecting the period dating.
\vspace{-10.5pt}

\paragraph{Knowledge Extraction Module (KEM).}
With this module, we first verify the influence of the shape information on the dating performance. After adding the shape head of the KEM to the baseline (2$\rightarrow$3), the dynasty dating accuracy of the network is reduced by $2.31\%$, and the period dating accuracy is reduced by $1.51\%$. After concatenating the extracted shape features with the period features and using them together for period learning (3$\rightarrow$4), the accuracy of the dynasty dating increases by $1.99\%$, and the accuracy of the period dating increases by $0.97\%$.

Second, we verify the influence of characteristic information on the dating performance. The addition of the characteristic head of the KEM (2$\rightarrow$6) improves the dating accuracy of the network by $0.38\%$ and $0.32\%$. After concatenating the extracted characteristic features with the period features and using them together for period learning (6$\rightarrow$7), the accuracy of the period dating decreases by $0.27\%$, whereas the accuracy of the dynasty dating improves by only $0.27\%$. When we simultaneously concatenate the shape and characteristic features into the features of the period for period learning (2$\rightarrow$9), but the accuracies of the dynasty dating and period dating decrease by $0.05\%$ and $0.43\%$, respectively.

These results demonstrate that concatenating the attribute information with the period information is inefficient. The external information is far from being fully utilized. In the following, we embed the shape and characteristic predictions into an archaeology knowledge-guided relation graph.

\begin{table*}[htbp]
\centering
\caption{Comparison of each method on the proposed datasets. Bold indicates the best results, and underlined values are the second best results. The single- and multi-granularity methods use single- and multi-granularity era labels as supervision, respectively. In addition, our method and $A^3$M~\cite{46} use additional attribute annotation information, and Part-based R-CNN~\cite{6} uses additional bounding box annotations.}
\label{tab3:Period ACC}
\resizebox{0.98\textwidth}{!}{  
\begin{tabular}{cc|c|cc|cc|ccc|ccc|ccc}
\hline\hline
\multicolumn{2}{c|}{}                                                                                           &                                 &                      &                                              & \multicolumn{2}{c|}{Shang}                  & \multicolumn{3}{c|}{Western Zhou}                                  & \multicolumn{3}{c|}{Spring and Autumn}                                                    & \multicolumn{3}{c}{Warring States}                                                        \\
\multicolumn{2}{c|}{\multirow{-2}{*}{Method}}                                                                   & \multirow{-2}{*}{w/ Attributes} & \multirow{-2}{*}{$OA$} & \multirow{-2}{*}{$AU(\overline{PRC})$}                    & Early                & Late                 & Early                & Mid                  & Late                 & Early                & Mid                  & Late                                        & Early                                       & Mid                  & Late                 \\ \hline\hline
\multicolumn{1}{c|}{}                                     & ConvNeXt~\cite{50}                  &                                 & 76.01                & {{\ul \textit{0.8397}}} & 73.04                & 82.31                & 75.21                & 82.56                & 80.00                & 77.44                & 64.17                & { {\ul \textit{67.76}}} & { {\ul \textit{56.90}}} & 41.18                & 63.37                \\ \cline{2-16} 
\multicolumn{1}{c|}{}                                     & Part-based R-CNN~\cite{6}           & BBox                    & 69.45                & 0.7796                                       & \textbf{92.59}       & {\ul \textit{84.47}} & 62.54                & \textbf{90.78}       & 67.74                & 62.03                & 62.79                & 46.89                                       & 38.46                                       & \textbf{73.33}       & \textbf{84.91}                \\ \cline{3-3}
\multicolumn{1}{c|}{}                                     & MCL~\cite{48}                       &                                 & 70.41                & 0.7742                                       & 79.55                & 78.56                & 66.85                & 79.58                & 76.28                & 71.57                & 54.23                & 66.02                                       & 35.42                                       & 31.18                & 60.11                \\ \cline{3-3}
\multicolumn{1}{c|}{}                                     & CrossX~\cite{30}                    &                                 & 70.54                & 0.7755                                       & 70.89                & 72.97                & 75.21                & 71.74                & 82.88                & 67.67                & 62.14                & 57.20                                       & 25.00                                       & 57.14                & 69.23                \\ \cline{3-3}
\multicolumn{1}{c|}{}                                     & BCNN~\cite{29}                      &                                 & 71.59                & 0.7402                                       & 71.15                & 80.48                & 70.92                & 76.82                & 78.60                & 73.14                & 57.53                & 59.00                                       & 0.00                                        & 0.00                 & 48.11                \\ \cline{3-3}
\multicolumn{1}{c|}{}                                     & NTS-Net~\cite{47}                   &                                 & 73.06                & 0.7890                                       & 71.15                & 80.93                & 67.28                & 79.17                & 78.15                & 81.56                & 58.24                & 60.48                                       & 54.17                                       & {\ul \textit{70.83}} & 67.78                \\ \cline{3-3}
\multicolumn{1}{c|}{}                                     & $A^3$M~\cite{46}                    & $\checkmark$                    & 75.12                & 0.8002                                       & 79.59                & 78.24                & 74.11                & 85.71                & 78.42                & 78.99                & 62.35                & 63.08                                       & 44.44                                       & 50.00                & 69.32                \\ \cline{3-3}
\multicolumn{1}{c|}{}                                     & SPS~\cite{31}                       &                                 & 76.94                & 0.8245                                       & 80.39                & 83.30                & 73.19                & {\ul \textit{86.51}} & {\ul \textit{85.61}} & 81.21                & 62.38                & 66.36                                       & 42.11                                       & 50.00                & 67.78                \\ \cline{3-3}
\multicolumn{1}{c|}{\multirow{-9}{*}{\rotatebox[origin=c]{90}{Single-Granularity}}} & P2PNet~\cite{45}                    &                                 & {\ul \textit{77.32}} & 0.8370                                       & 79.25                & 78.25                & \textbf{80.39}       & 78.80                & \textbf{88.37}       & \textbf{85.11}       & 64.52                & 67.24                                       & 50.00                                       & 52.78                & 71.59                \\ \hline
\multicolumn{1}{c|}{}                                     &                                                     &                                 & 84.85                & {\ul \textit{0.9125}}                        & \multicolumn{2}{c|}{{\ul \textit{84.37}}}   & \multicolumn{3}{c|}{84.09}                                         & \multicolumn{3}{c|}{{\ul \textit{87.25}}}                                                 & \multicolumn{3}{c}{84.97}                                                                 \\
\multicolumn{1}{c|}{}                                     & \multirow{-2}{*}{YourFL~\cite{1}}   & \multirow{-2}{*}{}              & 73.92                & 0.8019                                       & 79.25                & 80.20                & 69.38                & 82.21                & 83.59                & 82.14                & 56.38                & 62.90                                       & 53.85                                       & 51.72                & 60.67                \\ \cline{3-16} 
\multicolumn{1}{c|}{}                                     &                                                     &                                 & 84.43                & 0.8934                                       & \multicolumn{2}{c|}{82.07}                  & \multicolumn{3}{c|}{84.28}                                         & \multicolumn{3}{c|}{85.91}                                                                & \multicolumn{3}{c}{{\ul \textit{90.41}}}                                                  \\
\multicolumn{1}{c|}{}                                     & \multirow{-2}{*}{C-HMCNN~\cite{43}} & \multirow{-2}{*}{}              & 74.52                & 0.7766                                       & {\ul \textit{81.25}} & 79.29                & 70.04                & 85.57                & 81.95                & 75.00                & \textbf{67.47}       & 60.45                                       & \textbf{58.06}                              & 45.24                & {\ul \textit{77.46}}       \\ \cline{3-16} 
\multicolumn{1}{c|}{}                                     &                                                     &                                 & {\ul \textit{85.28}} & 0.9124                                       & \multicolumn{2}{c|}{81.08}                  & \multicolumn{3}{c|}{{\ul \textit{88.24}}}                          & \multicolumn{3}{c|}{85.98}                                                                & \multicolumn{3}{c}{85.71}                                                                 \\
\multicolumn{1}{c|}{}                                     & \multirow{-2}{*}{HRN~\cite{2}}      & \multirow{-2}{*}{}              & 75.81                & 0.8206                                       & 75.93                & 79.66                & 75.89                & 80.63                & 85.61                & 81.70                & 62.24                & 62.71                                       & 51.06                                       & 48.15                & 68.06                \\ \cline{3-16} 
\multicolumn{1}{c|}{}                                     &                                                     &                                 & \textbf{88.79}       & \textbf{0.9380}                              & \multicolumn{2}{c|}{\textbf{86.80}}         & \multicolumn{3}{c|}{\textbf{88.62}}                                & \multicolumn{3}{c|}{\textbf{91.57}}                                                       & \multicolumn{3}{c}{\textbf{90.45}}                                                        \\
\multicolumn{1}{c|}{\multirow{-8}{*}{\rotatebox[origin=c]{90}{Multi-Granularity}}}  & \multirow{-2}{*}{Ours}                              & \multirow{-2}{*}{$\checkmark$}  & \textbf{78.83}       & \textbf{0.8550}                              & 77.36                & \textbf{84.85}       & {\ul \textit{78.30}} & 81.28                & 85.31                & {\ul \textit{83.33}} & {\ul \textit{64.84}} & \textbf{68.85}                              & 45.95                                       & 53.13                & 75.31 \\ \hline
\end{tabular}
}
\end{table*}

\begin{table}[]
\centering
\caption{Comparison results of the dating performance showing the effect of the order of the shape and characteristic embedding.}
\label{tab:order}
\resizebox{.32\textwidth}{!}{  
\begin{tabular}{cl|cc}
\hline\hline
\multicolumn{2}{c|}{Embedding order}          & Dynasty $OA$     & Period $OA$      \\ \hline\hline
\multicolumn{2}{c|}{Era-Characteristic-Shape} & 86.37          & 76.13          \\ \hline
\multicolumn{2}{c|}{Era-Shape-Characteristic} & \textbf{88.79} & \textbf{78.83} \\ \hline
\end{tabular}
}
\end{table}

\vspace{-10.5pt}

\paragraph{Archaeology Knowledge Guided Relation Graph (AKG).}  
To make better use of the additional information, we build an AKG by adding the shape and characteristic predictions extracted from the KEM to the relation graph. Experiment results show that embedding the shape into the relation graph (2$\rightarrow$5) improves the accuracy of the dynasty dating by $0.86\%$ and the accuracy of the period dating by $0.81\%$. Embedding the characteristics into the relation graph (2$\rightarrow$8) improves the accuracy of dynasty dating by $0.87\%$ and accuracy of period dating by $0.93\%$. In the complete model, embedding both the shape and characteristics (2$\rightarrow$12) improves the accuracy of the dynasty dating by $1.94\%$ and the accuracy of the period dating by $1.78\%$.

It is worth noting that the embedding order of the shape and characteristics also has an impact on the dating performance. As illustrated in Equations (\ref{eq:Le-s}) and (\ref{eq:Le-a}), in our network, we use the Era-Shape-Characteristic order to incrementally learn different information. To verify the influence of the embedding order, we also tested the opposite case. In the Era-Characteristic-Shape learning order, when the model is unable to accurately determine the era of a sample based on its era feature, it first increases the influence of the characteristic features and finally considers the shape feature. The comparison results are shown in Table~\ref{tab:order}, where the embedding learning order of the Era-Characteristic-Shape is $2.42\%$ and $2.70\%$ lower than that of our applied order. This confirms that when learning labels with different distributions, the network needs to learn the labels progressively from weak to strong and from easy to complex. This also explains why the strong supervised method (Part-based R-CNN~\cite{6}) did not achieve excellent dating results in the following comparison experiment.
\begin{figure*}[htbp]
\centering
  \includegraphics[width=1.0\linewidth]{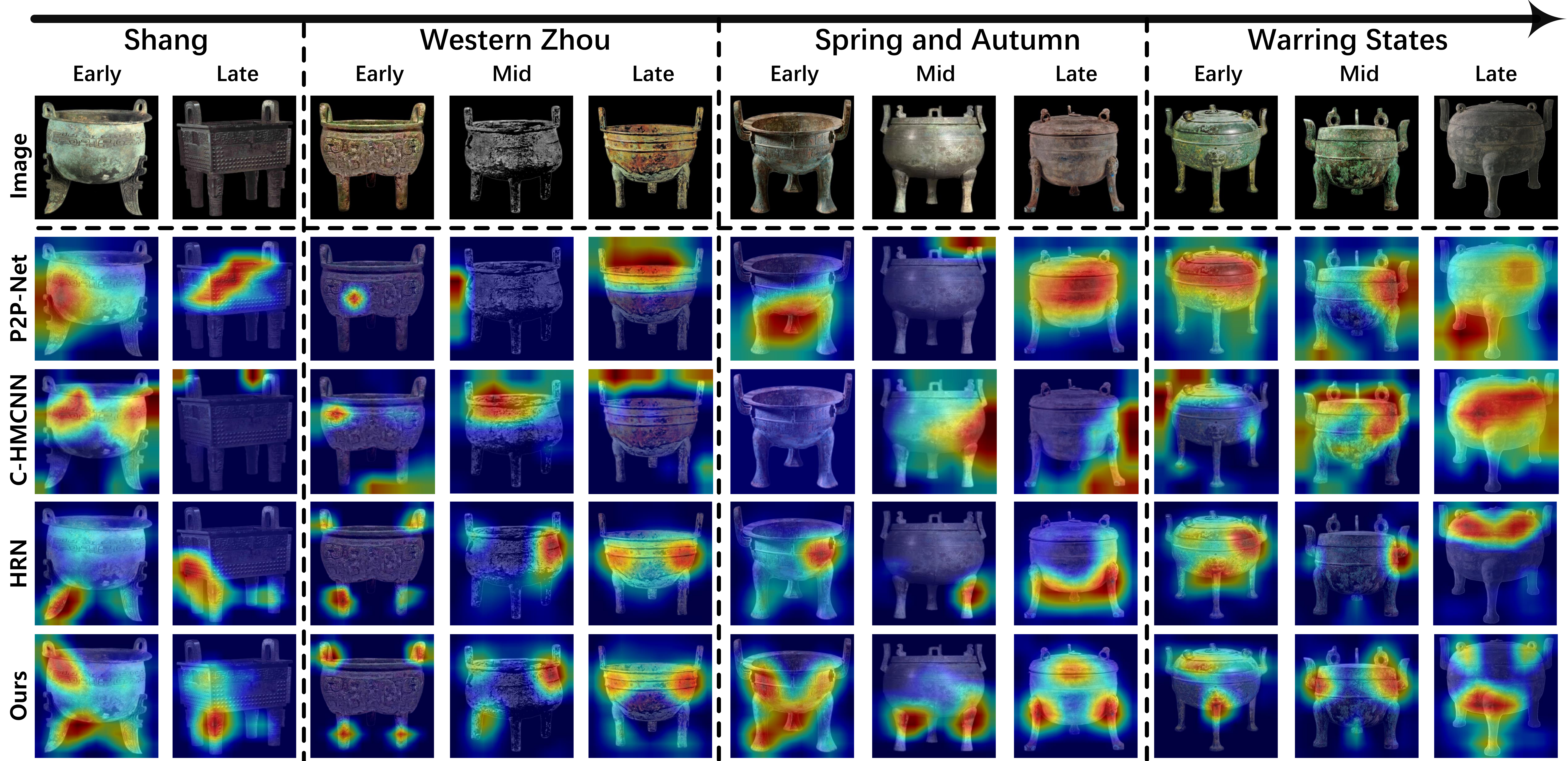}
  \caption{Gradient-weighted class activation map of different methods on 11 period test samples. Compared to the other methods, our network is more concentrated on the discriminative regions of a bronze ding and is able to capture its key characteristics.}
  \label{fig6:CAM}
\end{figure*}

\begin{figure}[htbp]
\centering
  \includegraphics[width=0.9\linewidth]{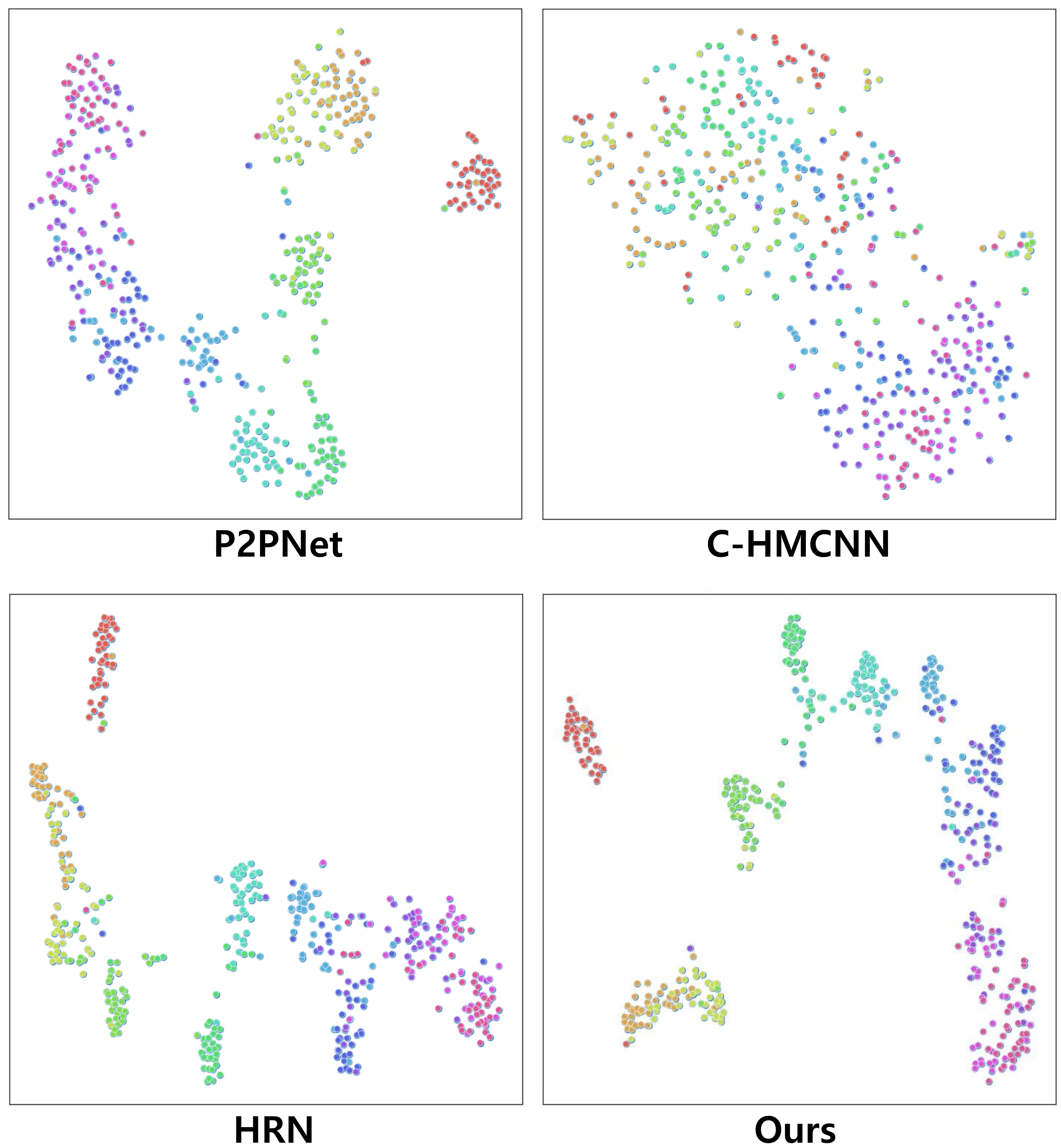}
  \caption{Visualization of learned representations of different methods on our dataset using T-SNE. Compared to the other methods, the decision boundaries of our proposed network become more separated.}
  \label{fig7:tsne}
\end{figure}

\subsection{Comparison with SOTA Methods}
\vspace{-4pt}
We compare our proposed network with other SOTA approaches under multi- and single-granularity settings. Under a multi-granularity setting, we train all multi-granularity methods with two-level labels of the bronze ding datasets. We report the $OA$ and $AU(\overline{PRC})$ results for each hierarchical level on the test set. Under a single-granularity setting, we train all single-granularity methods using fine-grained period labels of a bronze ding. Besides, to observe the classification performance of each dynasty and period independently, we calculate the precision of each approach on 4 coarse-grained dynasties as well as on 11 fine-grained periods recall is reported in our supplementary material. In addition to research related to fine-grained classification, we also compare our method with ConvNeXt~\cite{50}, which is an extremely popular approach in traditional classification tasks.

As Table~\ref{tab3:Period ACC} shows, using the same backbone ResNet50, our method outperforms the state-of-the-art 
single-granularity method P2PNet~\cite{45} on the bronze ding dataset benchmark by more than $1.51\%$ $OA$ and $0.018$ $AU(\overline{PRC})$ for period dating. And our method outperforms the state-of-the-art 
multi-granularity method HRN~\cite{2} (our baseline) on the bronze ding dataset benchmark by more than $3.51\%$ $OA$ and $0.0256$ $AU(\overline{PRC})$ in terms of dynasty dating, and by more than $3.02\%$ $OA$ and $0.0344$ $AU(\overline{PRC})$ for period dating. Furthermore, we achieve the best performance for all 4 coarse-grained dynasties and 3 out of 11 fine-grained periods for each independent era classification.

\vspace{-10.5pt}

\paragraph{Visualization.}

To demonstrate that our network can capture important regions of interest useful for bronze ding dating, we adopt Grad-CAM~\cite{52} for an intuitive visualization. For comparison, we also conducte the same visualization for the single-granularity method P2PNet~\cite{45} and the multi-granularity methods HRN~\cite{2} and C-HMCNN~\cite{43}, which also exhibit competitive performances. As shown in Figure~\ref{fig6:CAM}, our network is more concentrated within the discriminative regions of a bronze ding. 
Compared to the other methods, our network captures the key locations on a bronze ding when performing the dating, such as decorations and inscriptions.

In addition, as shown in Figure~\ref{fig7:tsne}, we draw t-SNE~\cite{53} scatter plots from the learned high-dimensional period features of our network and some other comparison methods. For better visualization, we randomly select 40 images for each period. From the t-SNE plots, we can clearly see that our network extracts more discriminative period representations of different images.

\vspace{-10.5pt}

\section{Conclusion and Further Work}
\vspace{-4pt}
In this study, we introduce a bronze ding dataset with rich archaeological labels. To address the challenges, we construct an end-to-end multihead network for predicting the era of bronze dings based on an AKG. Comprehensive experiments are conducted on our dataset, the results of which demonstrate the effectiveness of our proposed network in comparison to existing multi- and single-granularity FGVC methods. And, excitingly, our learning-based dating network achieve the same level of human experts. 

In a future study, we plan to collect and open up more types of Chinese bronze data to facilitate research through learning-based methods. We hope that our study will provide further contributions to both the archaeological and deep-learning communities.

\vspace{-10.5pt}

\paragraph{Acknowledgement.} This work is supported by the "Paleography and Chinese Civilization Inheritance and Development Program" Collaborative Innovation Platform (Grant No.G3829) and Jilin University (Grant No.419021421665 and No.419021422B08).




{\small
\bibliographystyle{ieee_fullname}
\bibliography{main}

\begin{thebibliography}{10}\itemsep=-1pt

\bibitem{64}
Yalong Bai, Yuxiang Chen, Wei Yu, Linfang Wang, and Wei Zhang.
\newblock Products-10k: A large-scale product recognition dataset.
\newblock {\em arXiv preprint arXiv:2008.10545}, 2020.

\bibitem{57}
Thomas Berg, Jiongxin Liu, Seung Woo~Lee, Michelle~L Alexander, David~W Jacobs,
  and Peter~N Belhumeur.
\newblock Birdsnap: Large-scale fine-grained visual categorization of birds.
\newblock In {\em Proceedings of the IEEE Conference on Computer Vision and
  Pattern Recognition}, pages 2011--2018, 2014.

\bibitem{60}
Lukas Bossard, Matthieu Guillaumin, and Luc~Van Gool.
\newblock Food-101--mining discriminative components with random forests.
\newblock In {\em European conference on computer vision}, pages 446--461.
  Springer, 2014.

\bibitem{39}
Ricardo Cerri, Rodrigo~C Barros, Andr{\'e}~C PLF~de Carvalho, and Yaochu Jin.
\newblock Reduction strategies for hierarchical multi-label classification in
  protein function prediction.
\newblock {\em BMC bioinformatics}, 17(1):1--24, 2016.

\bibitem{48}
Dongliang Chang, Yifeng Ding, Jiyang Xie, Ayan~Kumar Bhunia, Xiaoxu Li, Zhanyu
  Ma, Ming Wu, Jun Guo, and Yi-Zhe Song.
\newblock The devil is in the channels: Mutual-channel loss for fine-grained
  image classification.
\newblock {\em IEEE Transactions on Image Processing}, 29:4683--4695, 2020.

\bibitem{1}
Dongliang Chang, Kaiyue Pang, Yixiao Zheng, Zhanyu Ma, Yi-Zhe Song, and Jun
  Guo.
\newblock Your" flamingo" is my" bird": Fine-grained, or not.
\newblock In {\em Proceedings of the IEEE/CVF Conference on Computer Vision and
  Pattern Recognition}, pages 11476--11485, 2021.

\bibitem{2}
Jingzhou Chen, Peng Wang, Jian Liu, and Yuntao Qian.
\newblock Label relation graphs enhanced hierarchical residual network for
  hierarchical multi-granularity classification.
\newblock In {\em Proceedings of the IEEE/CVF Conference on Computer Vision and
  Pattern Recognition}, pages 4858--4867, 2022.

\bibitem{69}
Tianshui Chen, Liang Lin, Riquan Chen, Yang Wu, and Xiaonan Luo.
\newblock Knowledge-embedded representation learning for fine-grained image
  recognition.
\newblock {\em arXiv preprint arXiv:1807.00505}, 2018.

\bibitem{65}
Tianshui Chen, Wenxi Wu, Yuefang Gao, Le Dong, Xiaonan Luo, and Liang Lin.
\newblock Fine-grained representation learning and recognition by exploiting
  hierarchical semantic embedding.
\newblock In {\em Proceedings of the 26th ACM international conference on
  Multimedia}, pages 2023--2031, 2018.

\bibitem{74}
Yin Cui, Feng Zhou, Yuanqing Lin, and Serge Belongie.
\newblock Fine-grained categorization and dataset bootstrapping using deep
  metric learning with humans in the loop.
\newblock In {\em Proceedings of the IEEE conference on computer vision and
  pattern recognition}, pages 1153--1162, 2016.

\bibitem{20}
Jia Deng, Nan Ding, Yangqing Jia, Andrea Frome, Kevin Murphy, Samy Bengio, Yuan
  Li, Hartmut Neven, and Hartwig Adam.
\newblock Large-scale object classification using label relation graphs.
\newblock In {\em European conference on computer vision}, pages 48--64.
  Springer, 2014.

\bibitem{38}
Jia Deng, Jonathan Krause, Michael Stark, and Li Fei-Fei.
\newblock Leveraging the wisdom of the crowd for fine-grained recognition.
\newblock {\em IEEE transactions on pattern analysis and machine intelligence},
  38(4):666--676, 2015.

\bibitem{11}
Antonio Dom{\'e}nech-Carb{\'o}, MT Dom{\'e}nech-Carb{\'o}, Jorge
  Redondo-Marug{\'a}n, Laura Osete-Cortina, J Barrio, A Fuentes, MV
  Vivancos-Ram{\'o}n, W Al~Sekhaneh, B Mart{\'\i}nez, I
  Mart{\'\i}nez-L{\'a}zaro, et~al.
\newblock Electrochemical characterization and dating of archaeological leaded
  bronze objects using the voltammetry of immobilized particles.
\newblock {\em Archaeometry}, 60(2):308--324, 2018.

\bibitem{10}
Antonio Dom{\'e}nech-Carb{\'o}, Mar{\'\i}a~Teresa Dom{\'e}nech-Carb{\'o}, Sofia
  Capelo, Trinidad Pas{\'\i}es, and Isabel Mart{\'\i}nez-L{\'a}zaro.
\newblock Dating archaeological copper/bronze artifacts by using the
  voltammetry of microparticles.
\newblock {\em Angewandte Chemie}, 126(35):9416--9420, 2014.

\bibitem{66}
Kun Duan, Devi Parikh, David Crandall, and Kristen Grauman.
\newblock Discovering localized attributes for fine-grained recognition.
\newblock In {\em 2012 IEEE conference on computer vision and pattern
  recognition}, pages 3474--3481. IEEE, 2012.

\bibitem{33}
Abhimanyu Dubey, Otkrist Gupta, Pei Guo, Ramesh Raskar, Ryan Farrell, and
  Nikhil Naik.
\newblock Pairwise confusion for fine-grained visual classification.
\newblock In {\em Proceedings of the European conference on computer vision
  (ECCV)}, pages 70--86, 2018.

\bibitem{32}
Abhimanyu Dubey, Otkrist Gupta, Ramesh Raskar, and Nikhil Naik.
\newblock Maximum-entropy fine grained classification.
\newblock {\em Advances in neural information processing systems}, 31, 2018.

\bibitem{62}
Yuying Ge, Ruimao Zhang, Xiaogang Wang, Xiaoou Tang, and Ping Luo.
\newblock Deepfashion2: A versatile benchmark for detection, pose estimation,
  segmentation and re-identification of clothing images.
\newblock In {\em Proceedings of the IEEE/CVF conference on computer vision and
  pattern recognition}, pages 5337--5345, 2019.

\bibitem{43}
Eleonora Giunchiglia and Thomas Lukasiewicz.
\newblock Coherent hierarchical multi-label classification networks.
\newblock {\em Advances in Neural Information Processing Systems},
  33:9662--9673, 2020.

\bibitem{46}
Kai Han, Jianyuan Guo, Chao Zhang, and Mingjian Zhu.
\newblock Attribute-aware attention model for fine-grained representation
  learning.
\newblock In {\em Proceedings of the 26th ACM international conference on
  Multimedia}, pages 2040--2048, 2018.

\bibitem{21}
Jiayuan He, Qingting Zhu, Youguang Chen, and Fan Nie.
\newblock Bronze inscriptions classification algorithm on imbalanced dataset.
\newblock In {\em 2020 5th International Conference on Mechanical, Control and
  Computer Engineering (ICMCCE)}, pages 1715--1718. IEEE, 2020.

\bibitem{5}
Kaiming He, Xiangyu Zhang, Shaoqing Ren, and Jian Sun.
\newblock Deep residual learning for image recognition.
\newblock In {\em Proceedings of the IEEE conference on computer vision and
  pattern recognition}, pages 770--778, 2016.

\bibitem{72}
Xiangteng He and Yuxin Peng.
\newblock Fine-grained image classification via combining vision and language.
\newblock In {\em Proceedings of the IEEE Conference on Computer Vision and
  Pattern Recognition}, pages 5994--6002, 2017.

\bibitem{59}
Saihui Hou, Yushan Feng, and Zilei Wang.
\newblock Vegfru: A domain-specific dataset for fine-grained visual
  categorization.
\newblock In {\em Proceedings of the IEEE International Conference on Computer
  Vision}, pages 541--549, 2017.

\bibitem{31}
Shaoli Huang, Xinchao Wang, and Dacheng Tao.
\newblock Stochastic partial swap: Enhanced model generalization and
  interpretability for fine-grained recognition.
\newblock In {\em Proceedings of the IEEE/CVF International Conference on
  Computer Vision}, pages 620--629, 2021.

\bibitem{9}
Shaoli Huang, Zhe Xu, Dacheng Tao, and Ya Zhang.
\newblock Part-stacked cnn for fine-grained visual categorization.
\newblock In {\em Proceedings of the IEEE conference on computer vision and
  pattern recognition}, pages 1173--1182, 2016.

\bibitem{61}
Menglin Jia, Mengyun Shi, Mikhail Sirotenko, Yin Cui, Claire Cardie, Bharath
  Hariharan, Hartwig Adam, and Serge Belongie.
\newblock Fashionpedia: Ontology, segmentation, and an attribute localization
  dataset.
\newblock In {\em European conference on computer vision}, pages 316--332.
  Springer, 2020.

\bibitem{14}
Aditya Khosla, Nityananda Jayadevaprakash, Bangpeng Yao, and Fei-Fei Li.
\newblock Novel dataset for fine-grained image categorization: Stanford dogs.
\newblock In {\em Proc. CVPR workshop on fine-grained visual categorization
  (FGVC)}, volume~2. Citeseer, 2011.

\bibitem{58}
Jonathan Krause, Michael Stark, Jia Deng, and Li Fei-Fei.
\newblock 3d object representations for fine-grained categorization.
\newblock In {\em Proceedings of the IEEE international conference on computer
  vision workshops}, pages 554--561, 2013.

\bibitem{3}
Alex Krizhevsky, Ilya Sutskever, and Geoffrey~E Hinton.
\newblock Imagenet classification with deep convolutional neural networks.
\newblock {\em Communications of the ACM}, 60(6):84--90, 2017.

\bibitem{22}
Bo Kuang, Youguang Chen, and Bin Su.
\newblock Detecting for bronze inscriptions.
\newblock In {\em Proceedings of the 2020 4th International Conference on
  Electronic Information Technology and Computer Engineering}, pages 555--559,
  2020.

\bibitem{42}
Tsung-Yi Lin, Priya Goyal, Ross Girshick, Kaiming He, and Piotr Doll{\'a}r.
\newblock Focal loss for dense object detection.
\newblock In {\em Proceedings of the IEEE international conference on computer
  vision}, pages 2980--2988, 2017.

\bibitem{29}
Tsung-Yu Lin, Aruni RoyChowdhury, and Subhransu Maji.
\newblock Bilinear cnn models for fine-grained visual recognition.
\newblock In {\em Proceedings of the IEEE international conference on computer
  vision}, pages 1449--1457, 2015.

\bibitem{24}
He Ling, Zhao Qingrong, and Gao Min.
\newblock Characterization of corroded bronze ding from the yin ruins of china.
\newblock {\em Corrosion science}, 49(6):2534--2546, 2007.

\bibitem{49}
Xiong Liu.
\newblock Identification and collection of bronze dings of all ages.
\newblock {\em Oriental Collection}, (8):121--124, 2014.

\bibitem{68}
Xiao Liu, Jiang Wang, Shilei Wen, Errui Ding, and Yuanqing Lin.
\newblock Localizing by describing: Attribute-guided attention localization for
  fine-grained recognition.
\newblock In {\em Thirty-First AAAI Conference on Artificial Intelligence},
  2017.

\bibitem{41}
Ziwei Liu, Ping Luo, Shi Qiu, Xiaogang Wang, and Xiaoou Tang.
\newblock Large-scale fashion (deepfashion) database.
\newblock {\em Xiaoou TangMultimedia Laboratory, The Chinese University of Hong
  Kong, Category and Attribute Prediction Benchmark. https://url. kr/dfQWlV},
  2016.

\bibitem{50}
Zhuang Liu, Hanzi Mao, Chao-Yuan Wu, Christoph Feichtenhofer, Trevor Darrell,
  and Saining Xie.
\newblock A convnet for the 2020s.
\newblock In {\em Proceedings of the IEEE/CVF Conference on Computer Vision and
  Pattern Recognition}, pages 11976--11986, 2022.

\bibitem{27}
Ilya Loshchilov and Frank Hutter.
\newblock Sgdr: Stochastic gradient descent with warm restarts.
\newblock {\em arXiv preprint arXiv:1608.03983}, 2016.

\bibitem{30}
Wei Luo, Xitong Yang, Xianjie Mo, Yuheng Lu, Larry~S Davis, Jun Li, Jian Yang,
  and Ser-Nam Lim.
\newblock Cross-x learning for fine-grained visual categorization.
\newblock In {\em Proceedings of the IEEE/CVF international conference on
  computer vision}, pages 8242--8251, 2019.

\bibitem{13}
Subhransu Maji, Esa Rahtu, Juho Kannala, Matthew Blaschko, and Andrea Vedaldi.
\newblock Fine-grained visual classification of aircraft.
\newblock {\em arXiv preprint arXiv:1306.5151}, 2013.

\bibitem{40}
Ishan Misra, Abhinav Shrivastava, Abhinav Gupta, and Martial Hebert.
\newblock Cross-stitch networks for multi-task learning.
\newblock In {\em Proceedings of the IEEE conference on computer vision and
  pattern recognition}, pages 3994--4003, 2016.

\bibitem{55}
Maria-Elena Nilsback and Andrew Zisserman.
\newblock Automated flower classification over a large number of classes.
\newblock In {\em 2008 Sixth Indian Conference on Computer Vision, Graphics \&
  Image Processing}, pages 722--729. IEEE, 2008.

\bibitem{51}
Adam Paszke, Sam Gross, Soumith Chintala, Gregory Chanan, Edward Yang, Zachary
  DeVito, Zeming Lin, Alban Desmaison, Luca Antiga, and Adam Lerer.
\newblock Automatic differentiation in pytorch.
\newblock 2017.

\bibitem{37}
Scott Reed, Zeynep Akata, Honglak Lee, and Bernt Schiele.
\newblock Learning deep representations of fine-grained visual descriptions.
\newblock In {\em Proceedings of the IEEE conference on computer vision and
  pattern recognition}, pages 49--58, 2016.

\bibitem{52}
Ramprasaath~R Selvaraju, Michael Cogswell, Abhishek Das, Ramakrishna Vedantam,
  Devi Parikh, and Dhruv Batra.
\newblock Grad-cam: Visual explanations from deep networks via gradient-based
  localization.
\newblock In {\em Proceedings of the IEEE international conference on computer
  vision}, pages 618--626, 2017.

\bibitem{4}
Karen Simonyan and Andrew Zisserman.
\newblock Very deep convolutional networks for large-scale image recognition.
\newblock {\em arXiv preprint arXiv:1409.1556}, 2014.

\bibitem{35}
Ming Sun, Yuchen Yuan, Feng Zhou, and Errui Ding.
\newblock Multi-attention multi-class constraint for fine-grained image
  recognition.
\newblock In {\em Proceedings of the European Conference on Computer Vision
  (ECCV)}, pages 805--821, 2018.

\bibitem{36}
Xiaoxiao Sun, Liyi Chen, and Jufeng Yang.
\newblock Learning from web data using adversarial discriminative neural
  networks for fine-grained classification.
\newblock In {\em Proceedings of the AAAI Conference on Artificial
  Intelligence}, volume~33, pages 273--280, 2019.

\bibitem{53}
Laurens Van~der Maaten and Geoffrey Hinton.
\newblock Visualizing data using t-sne.
\newblock {\em Journal of machine learning research}, 9(11), 2008.

\bibitem{56}
Grant Van~Horn, Steve Branson, Ryan Farrell, Scott Haber, Jessie Barry, Panos
  Ipeirotis, Pietro Perona, and Serge Belongie.
\newblock Building a bird recognition app and large scale dataset with citizen
  scientists: The fine print in fine-grained dataset collection.
\newblock In {\em Proceedings of the IEEE Conference on Computer Vision and
  Pattern Recognition}, pages 595--604, 2015.

\bibitem{12}
Catherine Wah, Steve Branson, Peter Welinder, Pietro Perona, and Serge
  Belongie.
\newblock The caltech-ucsd birds-200-2011 dataset.
\newblock 2011.

\bibitem{28}
Shimin Wang, Gongrou Chen, and Changshou Zhang.
\newblock A study of the phasing and dating of western zhou bronzes.
\newblock {\em Artifacts}, (9):96--96, 2019.

\bibitem{23}
Yanjie Wang, Guofeng Wei, Qiang Li, Xiaoping Zheng, and Danchun Wang.
\newblock Provenance of zhou dynasty bronze vessels unearthed from zongyang
  county, anhui province, china: determined by lead isotopes and trace
  elements.
\newblock {\em Heritage Science}, 9(1):1--12, 2021.

\bibitem{63}
Xiu-Shen Wei, Quan Cui, Lei Yang, Peng Wang, and Lingqiao Liu.
\newblock Rpc: A large-scale retail product checkout dataset.
\newblock {\em arXiv preprint arXiv:1901.07249}, 2019.

\bibitem{44}
Xiu-Shen Wei, Yi-Zhe Song, Oisin Mac~Aodha, Jianxin Wu, Yuxin Peng, Jinhui
  Tang, Jian Yang, and Serge Belongie.
\newblock Fine-grained image analysis with deep learning: A survey.
\newblock {\em IEEE Transactions on Pattern Analysis and Machine Intelligence},
  2021.

\bibitem{8}
Xiu-Shen Wei, Chen-Wei Xie, Jianxin Wu, and Chunhua Shen.
\newblock Mask-cnn: Localizing parts and selecting descriptors for fine-grained
  bird species categorization.
\newblock {\em Pattern Recognition}, 76:704--714, 2018.

\bibitem{54}
Zhenfeng Wu.
\newblock A collection of shang and zhou bronze inscriptions and images.
\newblock {\em Shanghai Ancient Books Publishing House}, 2012.

\bibitem{73}
Huapeng Xu, Guilin Qi, Jingjing Li, Meng Wang, Kang Xu, and Huan Gao.
\newblock Fine-grained image classification by visual-semantic embedding.
\newblock In {\em IJCAI}, pages 1043--1049, 2018.

\bibitem{70}
Zhe Xu, Shaoli Huang, Ya Zhang, and Dacheng Tao.
\newblock Webly-supervised fine-grained visual categorization via deep domain
  adaptation.
\newblock {\em IEEE transactions on pattern analysis and machine intelligence},
  40(5):1100--1113, 2016.

\bibitem{45}
Xuhui Yang, Yaowei Wang, Ke Chen, Yong Xu, and Yonghong Tian.
\newblock Fine-grained object classification via self-supervised pose
  alignment.
\newblock In {\em Proceedings of the IEEE/CVF Conference on Computer Vision and
  Pattern Recognition}, pages 7399--7408, 2022.

\bibitem{47}
Ze Yang, Tiange Luo, Dong Wang, Zhiqiang Hu, Jun Gao, and Liwei Wang.
\newblock Learning to navigate for fine-grained classification.
\newblock In {\em Proceedings of the European Conference on Computer Vision
  (ECCV)}, pages 420--435, 2018.

\bibitem{7}
Han Zhang, Tao Xu, Mohamed Elhoseiny, Xiaolei Huang, Shaoting Zhang, Ahmed
  Elgammal, and Dimitris Metaxas.
\newblock Spda-cnn: Unifying semantic part detection and abstraction for
  fine-grained recognition.
\newblock In {\em Proceedings of the IEEE conference on computer vision and
  pattern recognition}, pages 1143--1152, 2016.

\bibitem{6}
Ning Zhang, Jeff Donahue, Ross Girshick, and Trevor Darrell.
\newblock Part-based r-cnns for fine-grained category detection.
\newblock In {\em European conference on computer vision}, pages 834--849.
  Springer, 2014.

\bibitem{67}
Ning Zhang, Ryan Farrell, Forrest Iandola, and Trevor Darrell.
\newblock Deformable part descriptors for fine-grained recognition and
  attribute prediction.
\newblock In {\em Proceedings of the IEEE International Conference on Computer
  Vision}, pages 729--736, 2013.

\bibitem{34}
Yifan Zhao, Ke Yan, Feiyue Huang, and Jia Li.
\newblock Graph-based high-order relation discovery for fine-grained
  recognition.
\newblock In {\em Proceedings of the IEEE/CVF Conference on Computer Vision and
  Pattern Recognition}, pages 15079--15088, 2021.

\bibitem{15}
Yi Zheng, Yunfeng Yan, and Donglian Qi.
\newblock Bronze inscription recognition with distribution calibration based on
  few-shot learning.
\newblock In {\em 4th International Conference on Informatics Engineering \&
  Information Science (ICIEIS2021)}, volume 12161, pages 150--158. SPIE, 2022.

\bibitem{71}
Bohan Zhuang, Lingqiao Liu, Yao Li, Chunhua Shen, and Ian Reid.
\newblock Attend in groups: a weakly-supervised deep learning framework for
  learning from web data.
\newblock In {\em Proceedings of the IEEE Conference on Computer Vision and
  Pattern Recognition}, pages 1878--1887, 2017.

\end{thebibliography}

}

\end{document}